\documentclass[journal]{IEEEtran}

\usepackage{cite}
\usepackage{amsmath,amssymb,amsfonts}
\usepackage{algorithmic}
\usepackage{graphicx}
\usepackage{textcomp}
\usepackage{xcolor}
\usepackage{wrapfig}
\usepackage[utf8]{inputenc}
\usepackage{graphicx}
\usepackage{cite}
\usepackage{indentfirst}
\usepackage[justification=centering]{caption}
\usepackage{amsmath}
\usepackage{lscape}
\usepackage{gensymb}
\usepackage{algorithm}
\usepackage{algorithmic}
\usepackage{mathtools}
\usepackage{amsfonts}
\usepackage{comment}
\usepackage{bm}
\usepackage{multirow}
\usepackage{marvosym}
\usepackage[dvipsnames]{xcolor}
\usepackage{multirow}
\usepackage{array}
\usepackage{makecell}
\usepackage{booktabs}
\usepackage[colorlinks=true, linkcolor=black, citecolor=black, urlcolor=black]{hyperref}
\def\BibTeX{{\rm B\kern-.05em{\sc i\kern-.025em b}\kern-.08em
    T\kern-.1667em\lower.7ex\hbox{E}\kern-.125emX}}

\begin{document}

\title{Towards Cognitive Defect Analysis in Active Infrared Thermography with Vision-Text Cues}

\author{Mohammed Salah,
        Eman Ouda,
        Giuseppe Dell'Avvocato,
        Fabrizio Sarasini,
        Ester D'Accardi,
        Jorge Dias,
        Davor Svetinovic,
        Stefano Sfarra,
        and Yusra Abdulrahman
\thanks{This research was funded by Khalifa University of Science and Technology through the [Advancing Non-Destructive Testing (NDT) through Innovative Integration of Infrared Thermography (IRT) and Emerging Technologies in Aerospace Applications] under Project ID: KU-[INT]-[FSU]-[2024]-[8474000660].}
\thanks{M. Salah, E. Ouda, and Y. Abdulrahman are with the Department of Aerospace Engineering, Khalifa University, Abu Dhabi, UAE. Y. Abdulrahman is also with the Advanced Research and Innovation Center (ARIC), Khalifa University, Abu Dhabi, UAE.

G. Dell'Avvocato ans S. Sfarra are with the Department of Industrial and Information Engineering and Economics (DIIIE), University of L'Aquila, Piazzale E. Pontieri 1, 67100 L'Aquila, Italy.

F. Sarasini is with the Department of Chemical Engineering Materials Environment \& UDR INSTM, Sapienza University of Rome, Rome, Italy.

E. D'Accardi is with the Department of Mechanics, Mathematics and Management (DMMM), Polytechnic University of Bari, Via 5 Orabona 4, 70125 Bari, Italy.

J. Dias is with the College of Computing and Mathematical Sciences, Khalifa University, Abu Dhabi, 127788, UAE.

D. Svetinovic is with the Department of Computer Science, Khalifa University of Science and Technology, Abu Dhabi, UAE.

Yusra Abdulrahman is the corresponding author (email: yusra.abdulrahman@ku.ac.ae).}}

\IEEEaftertitletext{\vspace{-3\baselineskip}}

\maketitle

\begin{abstract}
Active infrared thermography (AIRT) is currently witnessing a surge of artificial intelligence (AI) methodologies being deployed for automated subsurface defect analysis of high performance carbon fiber-reinforced polymers (CFRP). Deploying AI-based AIRT methodologies for inspecting CFRPs requires the creation of time consuming and expensive datasets of CFRP inspection sequences to train neural networks. To address this challenge, this work introduces a novel language-guided framework for cognitive defect analysis in CFRPs using AIRT and vision-language models (VLMs). Unlike conventional learning-based approaches, the proposed framework does not require developing training datasets for extensive training of defect detectors, instead it relies solely on pretrained multimodal VLM encoders coupled with a lightweight adapter to enable generative zero-shot understanding and localization of subsurface defects. By leveraging pretrained multimodal encoders, the proposed system enables generative zero-shot understanding of thermographic patterns and automatic detection of subsurface defects. Given the domain gap between thermographic data and natural images used to train VLMs, an AIRT-VLM Adapter is proposed to enhance the visibility of defects while aligning the thermographic domain with the learned representations of VLMs. The proposed framework is validated using three representative VLMs; specifically, GroundingDINO, Qwen-VL-Chat, and CogVLM. Validation is performed on 25 CFRP inspection sequences with impacts introduced at different energy levels, reflecting realistic defects encountered in industrial scenarios. Experimental results demonstrate that the AIRT-VLM adapter achieves signal-to-noise ratio (SNR) gains exceeding 10 dB compared with conventional thermographic dimensionality-reduction methods, while enabling zero-shot defect detection with intersection-over-union (IoU) values reaching approximately 70\%. These findings indicate that coupling pretrained VLMs with the proposed adapter enables reliable localization of subsurface CFRP defects without defect-specific training and extensive dataset preparation.
\end{abstract}
 

\section{Introduction}

Carbon fiber reinforced polymers (CFRPs) are valued in the aerospace industry for their exceptional strength-to-weight ratio, corrosion resistance, and fatigue performance, enabling lighter airframes, improved fuel efficiency, and extended service life. Modern transport aircraft now rely heavily on CFRPs in primary load-bearing structures such as fuselage skins, wings, spars, ribs, and tail sections, as well as in secondary components, including control surfaces, fairings, doors, nacelles, and interior panels \cite{frp_insight,material_performance}. However, during manufacturing and in service, CFRP parts can develop a variety of defects and damage modes, such as porosity, resin-rich areas, fiber waviness or wrinkling, matrix cracking, debonding, delaminations, and impact-induced barely visible impact damage, which degrade stiffness, strength, and fatigue life and may remain hidden from visual inspection \cite{overview_ndt}.

To assess material properties, structural integrity, and subsurface defects without causing damage, nondestructive testing (NDT) techniques are employed, particularly in safety-critical fields such as aerospace. Commonly used NDT methods for CFRP inspection include ultrasonic testing, radiographic inspection, and infrared thermography (IRT), all of which can detect hidden anomalies and subsurface damage \cite{Tai2025}. Among these techniques, IRT has emerged as a valuable tool for identifying both surface and subsurface defects by analyzing the thermal response of CFRP structures \cite{stft}.

IRT is based on monitoring the propagation of heat on the surface of a material, where disturbances in thermal patterns indicate the presence of internal anomalies \cite{yusra_depth}. In particular, active infrared thermography (AIRT) improves defect detectability by applying external thermal excitation, such as flash lamps, halogen heaters, or lasers, which increases the thermal contrast between sound and defective regions \cite{yusra_4}. As a result, AIRT is especially suitable for inspecting non-metallic and multilayered materials where conventional inspection methods may be limited. AIRT has been widely adopted in sectors such as aerospace, energy, and civil infrastructure, where large-scale and on-site inspections are often required. Furthermore, recent advances in artificial intelligence (AI) have enabled automated defect characterization in AIRT. AI-based pulsed thermography (PT) algorithms have been proposed for defect classification \cite{flexible_framework}, segmentation \cite{attention_unet, pt_dataset}, and depth estimation \cite{3d_cnn}. In addition, to accelerate thermographic inspection processes and enable coverage of large structures, robotic and line-scan thermography systems have been introduced \cite{drone_sites}.

Although AI methodologies are currently being investigated in AIRT, a challenge in AI-based AIRT is the scarcity of datasets and the need to prepare costly datasets for training AI models for defect analysis. Vision–Language Models (VLMs) offer a promising paradigm for zero-shot defect detection in AIRT; however, current thermographic representations produced by conventional dimensionality-reduction techniques are not designed to generate image representations aligned with the natural-image domain of foundation VLMs, limiting their direct applicability for zero-shot reasoning. To address these challenges, this work proposes a zero-shot cognitive framework for defect analysis in AIRT using vision and text cues, leveraging the strong reasoning capabilities of pretrained multimodal VLMs. Specifically, an AIRT–VLM adapter is introduced to transform thermographic information into VLM-compatible representations, enabling off-the-shelf VLMs to perform zero-shot subsurface defect localization without thermography-specific training or large annotated datasets. As such, the focus of this work is methodological and AI-driven, applying multimodal VLMs to thermographic data for zero-shot defect localization, rather than advancing the physical modeling of infrared thermography.

The key contributions of this work are as follows:
\begin{enumerate}
    \item A novel zero-shot cognitive defect analysis framework for CFRP components is introduced, addressing the challenge of preparing time-consuming, costly datasets for AI-based thermographic inspection.
    \item The AIRT-VLM Adapter is proposed to bridge the domain gap between thermographic data and natural image distributions used in pretrained Vision–Language Models, enhancing the visibility of the defect and representation alignment.
    \item The proposed framework is tested to detect impact damage at different energy levels. The results show that VLMs coupled with the AIRT-VLM adapter enable reliable grounding of subsurface defects. 
\end{enumerate}

\subsection{Related Work}

Compared to earlier NDT techniques, such as radiography, eddy current testing, and ultrasonic testing, AIRT offers higher efficiency, faster evaluation, and fully non-contact inspection \cite{yusra_3}. This growing interest has spurred extensive research into learning-based models for defect detection, including adaptations of Faster R-CNN and YOLOv5 for IRT data \cite{flexible_framework}, as well as ConvLSTM architectures to better capture temporal dependencies \cite{cnnlstm}.
For defect segmentation, various neural network architectures have been explored, such as U-Net variants for composites and forged components \cite{zhang2025automatic, mueller2022defect}, and ConvLSTM-based models for 3D defect depth reconstruction. However, many approaches rely mainly on spatial information and underutilize temporal dynamics in thermographic sequences. To address this limitation, 3D CNNs have been introduced to explicitly model spatiotemporal features and improve subsurface defect segmentation \cite{3d_cnn}. However, the aforementioned models are supervised and their performance depends heavily on the availability of large, carefully annotated thermographic datasets, whose acquisition and labeling are costly, time-consuming, and often limited in generalizability across varying inspection conditions.


In these methods, AIRT dimensionality reduction techniques are primarily used as preprocessing steps to generate compact thermographic representations suitable for deep neural network input, since raw inspection sequences are highly dimensional and computationally expensive to process. AIRT dimensionality reduction is used to generate compact thermographic representations suitable for deep neural network input, as raw inspection sequences are highly dimensional and computationally expensive to process. Such techniques serve two main purposes: compressing thousands of frames into low-dimensional feature vectors and enhancing defect visibility by improving the typically low signal-to-noise ratio of raw thermal images, which otherwise degrades defect characterization performance \cite{salah_fusion}. Common dimensionality reduction methods include Thermal Signal Reconstruction (TSR) \cite{tsr_2}, Pulsed Phase Thermography (PPT) \cite{ppt_1}, and Principal Component Analysis (PCA) \cite{pct}. Physics-informed PCA variants have been proposed to improve AIRT analysis \cite{sparse}, and such representations are widely used as preprocessing in AI-based pipelines, including multimodal fusion approaches \cite{salah_fusion}. More recently, data-driven autoencoders, particularly CNN-based models, have been employed to learn compact latent features that capture nonlinear spatial and temporal patterns in thermographic data \cite{dat, 1d_cnn, yusra_taguchi, salah_fusion}. 

Therefore, defect detection is feasible with learning-based AIRT, which follows the traditional deep neural network training pipeline: inspection sequences are first collected as training data, then processed through dimensionality reduction to obtain compact thermographic representations, and finally used to train networks that are later deployed for downstream tasks. However, this pipeline suffers from two major drawbacks: preparing AIRT datasets for neural network training is costly and time-consuming, and traditional dimensionality-reduction methods do not guarantee a unified, image-like representation suitable for foundation-level, generalizable models capable of zero-shot defect detection. Moreover, the resulting thermographic representations are not inherently aligned with the natural-image domain on which vision–language models (VLMs) are pretrained, further limiting their direct applicability to zero-shot cognitive analysis. Hence, instead of relying on extensive data preparation and hand-crafted dimensionality-reduction methods, this work proposes a zero-shot cognitive framework for defect analysis in AIRT using vision–text cues, without the need for large-scale data collection, preparation, or defect-specific training.

The remaining sections of this paper are organized as follows. Section \ref{sec:methods} outlines the sample preparation and methodology. Section \ref{sec:exps} presents the experimental validations of the proposed methodology. Finally, section \ref{sec:conc} presents conclusions, findings, and future aspects of the proposed framework.

\section{Materials and Methods} \label{sec:methods}
\subsection{Specimens and Data Collection}

\subsubsection{Specimens}
Two types of additively manufactured carbon fiber reinforced polymer specimens were investigated: a PEKK matrix reinforced with continuous carbon fibers (PEKK-CF) and a PA12 matrix reinforced with short carbon fibers (PA12-CF). The PEKK-CF material consisted of a cross-ply lay-up obtained by stacking six 0°/90° plies containing continuous fibers, separated by intermediate layers of PEKK with short fiber reinforcement. Subsequently, it was consolidated by hot pressing at approximately 450 °C to achieve a symmetric laminate and satisfactory interlaminar bonding. The PA12-CF specimens were produced by fused filament fabrication through alternating 0°/90° deposition, resulting in a layer-by-layer architecture with a relatively homogeneous in-plane fiber distribution and reduced through-thickness continuity as described in previous work \cite{Moskovchenko2025}.

\begin{figure}[t]
    \centering
    \includegraphics[width=0.85\linewidth]{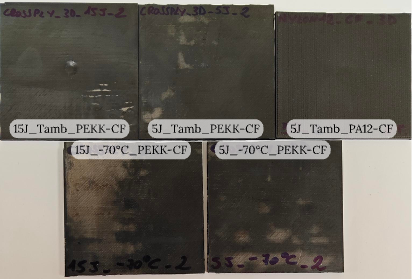}
    \caption{Front-side view of the impacted specimens, subjected to low-velocity impact at 5 J and 15 J.}
    \label{fig: Gpic1}
\end{figure}

All specimens had nominal dimensions of approximately 75 × 75 $mm^2$ and a thickness of about 4 $mm$. Low-velocity impact tests were carried out on the central area of each specimen using a drop-tower configuration. In the present work, specimens impacted at 5 J and 15 J were analyzed, with two PEKK samples tested at $-70^\circ\mathrm{C}$ and the remaining ones impacted at room temperature. The specimens were inspected with the surface as it is, i.e., in the same condition shown in the Fig. \ref{fig: Gpic1}. The visible indentation marks correspond to the main impact location and were used as a reference when defining the inspection area in the thermographic tests.

In the PEKK-CF laminate, the cross-ply arrangement of continuous fibers produces direction-dependent thermal transport within the plane. In contrast, in the PA12-CF specimens, the use of short fibers yields an almost isotropic response in the x–y plane and comparatively low diffusivity \cite{dell2025}.

\subsubsection{Data Collection}
\begin{figure}[b]
    \centering
    \includegraphics[width=0.7\linewidth]{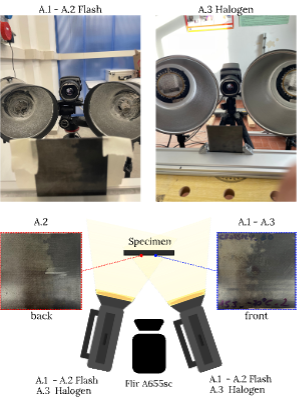}
    \caption{Top: Front-side heating with pulsed flash lamps (A.1-A.2) and front-side long-pulse halogen heating (A.3). Bottom: Schematic representation of inspection setup.}
    \label{fig: Gpic2}
\end{figure}

\begin{figure}[t]
    \centering
    \includegraphics[width=0.7\linewidth]{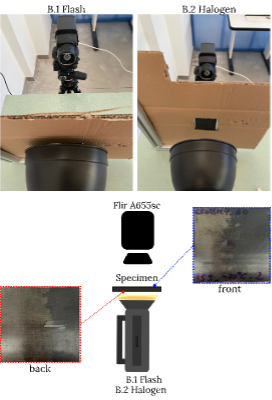}
    \caption{Top: Backside heating with flash (B.1) and halogen lamp (B.2). Bottom: Schematic representation of transmission inspection setup.}
    \label{fig: Gpic3}
\end{figure}

The specimens were tested under different heating configurations involving both reflection and transmission modes. The reflection modes involve inspection from the front side with pulsed flash heating and long-pulse heating,  and the back side with pulsed flash heating using halogen lamps (see Fig. \ref{fig: Gpic2}). On the other hand, the transmission mode involves inspection from the front side with pulsed flash heating and long-pulse heating using halogen lamps, see Fig.~\ref{fig: Gpic3}.

\begin{figure*}[!ht]
    \centering
    \includegraphics[keepaspectratio=true,scale=0.45, width=\linewidth]{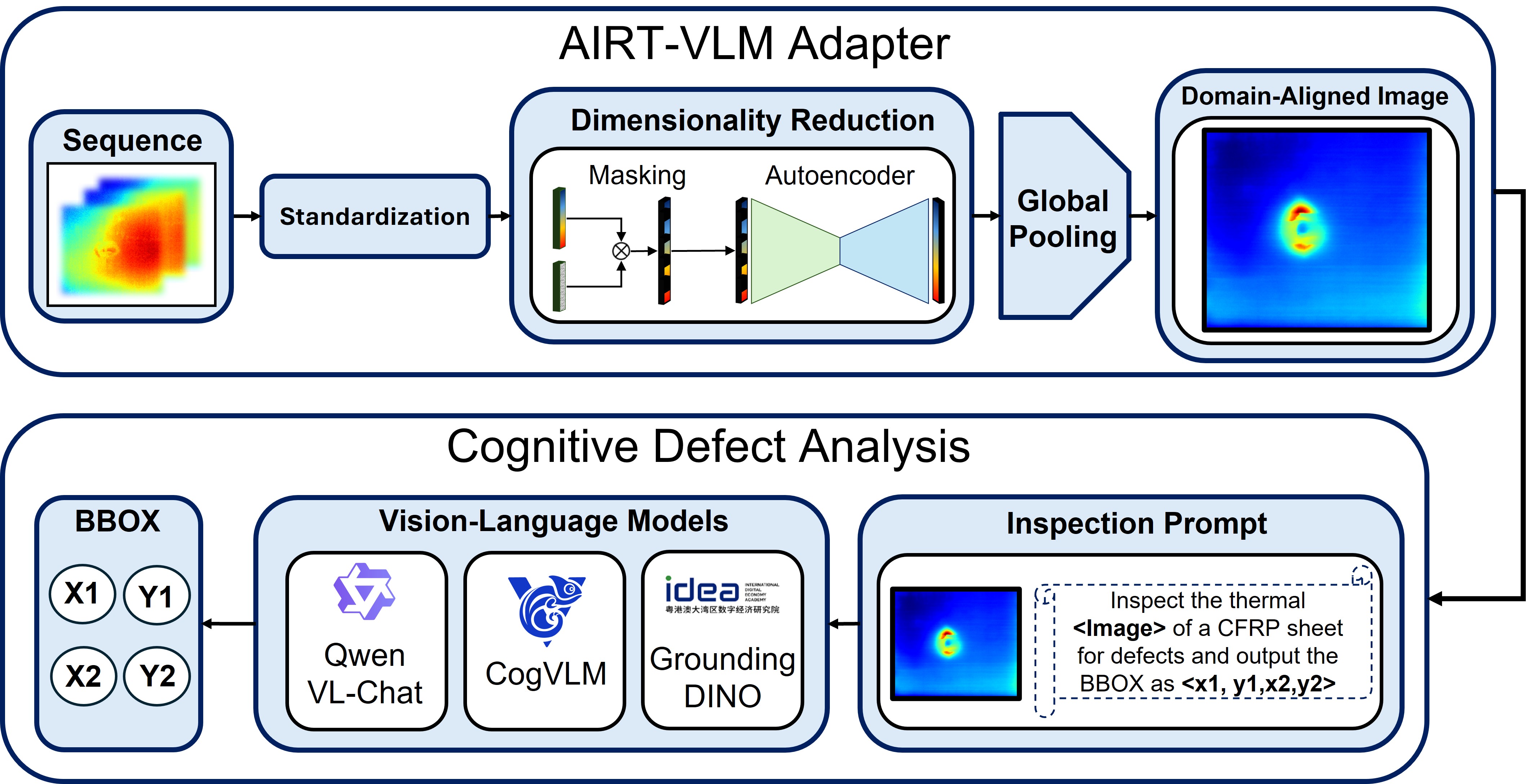}
    \caption{Overview of the defect analysis framework. The inspection sequence is preprocessed to generate a compact single image representation similar to the VLMs pretraining domain. Consequently, a VLM generates bounding boxes for defects.}
    \label{fig:framework}
\end{figure*}

In configurations A.1 and A.2, the specimen surface was heated using two flash lamps with a nominal energy of 3 kJ each. The lamps were placed laterally with respect to the infrared camera, which was positioned orthogonally to the inspected surface. This central positioning was selected to minimize dimensional errors due to perspective effects. The lamp arrangement, schematically illustrated in Fig. \ref{fig: Gpic2}, was defined to ensure sufficiently uniform heating of the specimen surface and to limit non-uniform heating artifacts. The surface was heated for approximately 4 ms, which, considering specimen thickness and previously estimated thermophysical properties \cite{dell2025}, is short enough to satisfy the impulsive heating assumption and justify the Dirac pulse approximation. The difference between A.1 and A.2 lies solely in the inspected surface: A.1 corresponds to the front surface (the impacted side), whereas A.2 corresponds to the rear surface. Configuration A.3 employed the same optical arrangement, i.e., two halogen lamps positioned laterally to the infrared camera and oriented as shown in Fig. \ref{fig: Gpic2}. In this case, two halogen sources with a nominal power of 650 W each were used. The surface was heated for a total duration of 20 s. Only the front surface was inspected in this configuration.

In configuration B.1, Fig. \ref{fig: Gpic3}, a single flash lamp with a nominal energy of 3 kJ was positioned at the minimum possible distance from the back surface of the specimen to maximize energy density transfer. To shield the infrared sensor and prevent pixel saturation from direct exposure to the light source, a cardboard frame was placed between the camera and the specimen. On the other hand, configuration B.2 used the same geometry adopted in B.1, with the only difference being the replacement of the flash source with a halogen lamp of 650 W nominal power. The back surface was heated for a duration of 40 s. Surface temperature evolution was recorded for all configurations using a FLIR A655sc microbolometric infrared camera operating in the 7–14 µm spectral range, with an acquisition frequency of 50 Hz. The resulting spatial resolution for reflection configuration A is 0.22 mm/pixel and for transmission configuration B is 0.19 mm/pixel.

\subsection{Methodology}
An overview of the proposed framework is shown in Fig.~\ref{fig:framework}. The input inspection sequence is first standardized to normalize its temporal and spatial dynamics across specimens. To interface AIRT data with VLMs, which are pretrained on natural RGB imagery, the AIRT-VLM adapter is devised as a dimensionality reduction module designed to compress the full thermal sequence into a single image representation with a denoised defect signal. The AIRT-VLM adapter is a masked autoencoder that extracts the dominant spatiotemporal features associated with subsurface defects. The autoencoder generates $l$ latent images that are pooled into a 2-D, single domain-aligned, thermal image that preserves defect visibility while remaining semantically closer to the distribution of images seen by VLMs during pre-training. Note that the domain-aligned image representation does not retain the full spatiotemporal physical information of the original AIRT sequence, but is instead optimized to enhance defect saliency and improve zero-shot localization performance when interfaced with VLMs. Accordingly, the domain-aligned image is subsequently fed into a VLM to generate a prediction of the bounding box $(x_1, y_1, x_2, y_2)$. Through this two-stage pipeline, a domain-adaptive reduction followed by VLM-driven reasoning, the framework enables a zero-shot VLM-compatible inspection process capable of detecting and localizing defects directly from thermographic sequences.

\subsubsection{AIRT-VLM Adapter} Let $\mathbf{S} = \{ I_{k} \}_{1}^{N_{t}}$, of shape $(N_{t}, N_{y}, N_{x})$, be the 3D matrix representing the inspection sequence, where $I_{k}$ is a thermogram timestamped at $k = 1, 2, \dots, N_{t}$, $N_{y}$ is the image height, and $N_{x}$ is its width. $\mathbf{S}$ is reshaped to $(N_{t}, N_{y} \times N_{x})$ by a raster-like operation and standardized by $\mathbf{\hat{S}} = \mathbf{S} - \mu_{k}$, where

\begin{equation}
    \mu_{k} = \frac{1}{N_{t}} \sum_{k=1}^{N_{t}} S^{(k)},
\end{equation}

\noindent and $\mathbf{\hat{S}} = \{ S^{(n)} \}_{1}^{N_{x}\times N_{y}}$ is a matrix consisting of the centered pixel-wise thermal responses. This operation is essential to ensure that the subsequent dimensionality reduction module focuses on the relative thermal variations induced by defects rather than absolute temperature offsets, thereby enhancing the discriminability of subtle defect signatures across different specimens and acquisition conditions. Afterwards, the $\mathbf{\hat{S}}$ is used to train the masked autoencoder to generate the compact latent representation.

The architecture of the autoencoder is the AIRT-Masked-CAAE, outlined in \cite{masked_ae}, leveraging the masked sequence autoencoding strategy for fast training. To ensure that the network avoids learning trivial, identity reconstruction, and focus on input signal features, $S^{(n)}$ is subjected to a binary masking operation with additive Gaussian noise, yielding a corrupted sequence, $\hat{S}^{(n)}$ by

\begin{equation}
    \hat{S}^{(n)} = M \odot S^{(n)} + \mathcal{N}(0,\sigma^2),
\end{equation}

\noindent where $M$ is a 1-D binary mask indicating visible ($1$) and masked ($0$) patches, and $\mathcal{N}(0,\sigma^2)$ represents additive Gaussian noise with zero mean and variance $\sigma^2$. Consequently, $\hat{S}^{(n)}$ is passed through the encoder, $f(\cdot)$, as

\begin{equation}
    \mathbf{z}_n = f_{\boldsymbol{\theta}}(\hat{S}^{(n)}),
\end{equation}

\noindent generating the latent representation $\mathbf{z}_{n}$. The decoder, $g(\cdot)$ aims to reconstruct the original, uncorrupted signal $\tilde{S}^{(n)}$ as

\begin{equation}
    \tilde{S}^{(n)} = g_{\boldsymbol{\phi}}(\mathbf{z}_n).
\end{equation}

\noindent The reconstruction strategy serves as guidance to train the network as

\begin{equation}
\mathcal{L} =
\frac{1}{N} \sum_{i=1}^{N} \| \tilde{S}_{i}^{(n)} - S_{i}^{(n)} \|_2^2.
\label{eq:recon_loss}
\end{equation}

\noindent Note that the training loss defined in \cite{pca_guided_ae} is the combined reconstruction-knowledge distillation loss to generate a structured latent space. For this work, we opt for a reconstruction focused loss function as it is aimed to generate a single domain-aligned image representation. In addition, the AIRT-Masked-CAAE is trained on the inspection sequence, compressed to a latent size of $l=10$, which is then pooled to generate the domain-aligned image. During training, optimization is performed using the Adam optimizer with a fixed learning rate of $10^{-3}$, a batch size of $32$, and training is conducted for $100$ epochs.

After online training, each $z^{(n)} \in \mathbf{z}_{n}$ is utilized as a pixel value, and its length is the number of channels, formulating $\mathbf{T}=\{T_{1}, T_{2}, \cdots, T_{l}\}$, where $l$ is the latent vector size. Thus, from all $K$ frames in the original image sequence, now the autoencoder latent images are only a few $l$ images. After obtaining the latent representation set $\mathbf{T}=\{T_{1}, T_{2}, \dots, T_{l}\}$ from the autoencoder, a global average pooling operation is applied to aggregate the multi-channel latent features into a single enhanced thermogram by
\begin{equation}
I_{\text{VLM}} = \text{Pool}\left(\mathbf{T}\right) = \frac{1}{l}\sum_{i=1}^{l} T_{i},
\end{equation}

\noindent where $I_{\text{VLM}}$ represents the high-SNR, domain-aligned thermal image, where the higher the contrast and SNR of $I_{\text{VLM}}$, the higher the detection accuracy and lower the uncertainty of the VLMs generated predictions. The domain-aligned image, $I_{\text{VLM}}$ is then fed to VLMs for generating the bounding box location of the subsurface defects.

\subsubsection{Cognitive Defect Analysis} VLMs possess multimodal reasoning capabilities of large pretrained models to detect and localize defects in a zero-shot manner. Given that $I_{\text{VLM}}$ is both high in contrast and semantically aligned to the statistics of natural images, it can be reliably interpreted by VLMs conditioned through natural-language prompts. The VLMs output the bounding box of the defect, identifying its location in the domain-aligned image. Let $\mathcal{P}$ denote the textual inspection instruction provided to the model, and let $\Phi(\cdot)$ represent a generic VLM composed of a visual encoder, a text encoder, and a multimodal fusion module. The VLM receives the paired input $(I_{\text{VLM}}, \mathcal{P})$ and produces a bounding-box estimate through
\begin{equation}
\mathbf{b} = \Phi\big(I_{\text{VLM}}, \mathcal{P}\big), \qquad 
\mathbf{b} = \langle x_{1}, y_{1}, x_{2}, y_{2} \rangle,
\label{eq:vlm_forward}
\end{equation}
where $\mathbf{b}$ contains the predicted defect localization coordinates. The visual encoder $f_{\text{vis}}(\cdot)$ extracts a feature representation of the thermal image,
\begin{equation}
\mathbf{v} = f_{\text{vis}}(I_{\text{VLM}}),
\label{eq:vis_enc}
\end{equation}
while the text encoder $f_{\text{text}}(\cdot)$ embeds the tokenized inspection prompt,
\begin{equation}
\mathbf{t} = f_{\text{text}}(\mathcal{P}).
\label{eq:text_enc}
\end{equation}
These unimodal representations are combined by a generic multimodal fusion operator $\mathcal{F}(\cdot, \cdot)$ that aligns the semantic information in the prompt with the visual content of the thermal image as
\begin{equation}
\mathbf{u} = \mathcal{F}(\mathbf{v}, \mathbf{t}).
\label{eq:fusion}
\end{equation}
The fused representation $\mathbf{u}$ captures the joint visual–linguistic understanding required to infer the presence and extent of subsurface defects. A prediction head $\psi(\cdot)$ operating on $\mathbf{u}$ yields the bounding-box estimate by
\begin{equation}
\mathbf{b} = \psi(\mathbf{u}).
\label{eq:generic_reg}
\end{equation}
This formulation is intentionally model-agnostic, allowing the VLM to be instantiated by any multimodal architecture, including those based on transformer cross-attention, region–text matching, or grounding-based detection heads. Regardless of the specific architecture, the VLM associates the high-SNR defect structures in $I_{\text{VLM}}$ with the semantic concept of a “defect” as defined in the inspection prompt. In this work, the prompt $\mathcal{P}$ takes the form
\begin{quote}
\texttt{Inspect the thermal image of a CFRP sheet and output the defect bounding box as $\langle x_{1}, y_{1}, x_{2}, y_{2} \rangle$.}
\end{quote}
This standardized instruction constrains the model’s output format and reduces ambiguity. Importantly, the VLM-driven analysis requires no thermal-domain fine-tuning or labeled thermographic data, as the reasoning capability emerges from the model’s large-scale multimodal pretraining. Consequently, the proposed approach enables flexible, data-efficient, and generalizable defect localization directly from pulse-thermography sequences through the AIRT-VLM adapter.

\begin{table*}[!ht]
\centering
\caption{Quantified contrast and SNR for the AIRT-VLM adapter representation benchmarked against state-of-the-art thermography dimensionality reduction methods, TSR, PCA, DAT \cite{dat}, 1D-DCAE-AIRT \cite{1d_cnn}, and C-AET \cite{constrained_ae}, under ambient and low-temperature ($-70^\circ$C) conditions.}
\label{tab:contrast_snr}
\resizebox{\textwidth}{!}{%
\begin{tabular}{l l c l c c c c c c c}
\toprule
\textbf{Defect Class} & \textbf{Temp.} & \textbf{Samples} & \textbf{Metric} & \textbf{Raw} & \textbf{TSR} & \textbf{PCA} & \textbf{DAT} & \textbf{1D-DCAE-AIRT} & \textbf{C-AET} & \textbf{Ours} \\
\midrule

\multirow{4}{*}{\textbf{5 J}}
& \multirow{2}{*}{Ambient}
& \multirow{2}{*}{6}
& Contrast & 0.207 & 0.241 & 0.302 & 0.366 & 0.383 & 0.361 & \textbf{0.478} \\
&&& SNR (dB) & 21.75 & 24.50 & 30.71 & 33.48 & 35.83 & 34.11 & \textbf{42.18} \\

& \multirow{2}{*}{$-70^\circ$C}
& \multirow{2}{*}{7}
& Contrast & 0.198 & 0.229 & 0.289 & 0.351 & 0.366 & 0.344 & \textbf{0.456} \\
&&& SNR (dB) & 20.90 & 23.62 & 29.40 & 32.05 & 34.21 & 32.88 & \textbf{40.27} \\

\addlinespace

\multirow{4}{*}{\textbf{15 J}}
& \multirow{2}{*}{Ambient}
& \multirow{2}{*}{7}
& Contrast & 0.227 & 0.287 & 0.387 & 0.395 & 0.436 & 0.387 & \textbf{0.534} \\
&&& SNR (dB) & 22.11 & 24.74 & 32.95 & 38.29 & 38.37 & 36.58 & \textbf{43.19} \\

& \multirow{2}{*}{$-70^\circ$C}
& \multirow{2}{*}{6}
& Contrast & 0.216 & 0.271 & 0.366 & 0.378 & 0.415 & 0.369 & \textbf{0.508} \\
&&& SNR (dB) & 21.08 & 23.90 & 31.42 & 36.10 & 36.84 & 35.02 & \textbf{41.36} \\

\bottomrule
\end{tabular}%
}
\end{table*}

\section{Experiments} \label{sec:exps}
\subsection{Experimental Evaluation}
The proposed framework is tested with three VLMs coupled with the proposed AIRT-VLM adapter; namely, CogVLM \cite{cogvlm}, Qwen-VL Chat \cite{qwen}, and GroundingDINO \cite{grounding_dino}. The experimental validation follows by evaluating the efficacy of the AIRT-VLM adapter in enhancing the defect signal and clarity in terms of contrast and SNR. Each is calculated as 

\begin{equation}
    \text{Contrast} = 
    \frac{\left| \left(\frac{1}{N}\sum_{p=1}^{N} Y_{d}(p)\right) - \left(\frac{1}{M}\sum_{q=1}^{M} Y_{s}(q)\right) \right|}
    {\left(\frac{1}{N}\sum_{p=1}^{N} Y_{d}(p)\right) + \left(\frac{1}{M}\sum_{q=1}^{M} Y_{s}(q)\right)},
\end{equation}

\begin{equation}
    \text{SNR} = 
    \frac{\left| \left(\frac{1}{N}\sum_{p=1}^{N} Y_{d}(p)\right) - \left(\frac{1}{M}\sum_{q=1}^{M} Y_{s}(q)\right) \right|}
    {\sigma_{s}},
\end{equation}

\noindent where $N$ denotes the total number of pixels in the defective region $Y_{d}$, with $Y_{d}(p)$ representing the $p^{\text{th}}$ pixel intensity in that region. $M$ refers to the number of pixels in the sound region $Y_{s}$, with $Y_{s}(q)$ being the $q^{\text{th}}$ pixel intensity, while $\sigma_{s}$ corresponds to the standard deviation of pixel values in the sound region $Y_{s}$. Note that the sound ROIs were obtained manually, while the ROI of the defective area is obtained from the labels. This evaluation is discussed in Section~\ref{subsec:adapter_eval}. In addition to signal enhancement evaluation, the more important task is assessing the defect-detection performance of the VLMs, as this represents the main objective and motivation of this work. The evaluation is performed using two metrics: the Intersection-over-Union (IoU) and the normalized center distance (NCD), defined respectively as
\begin{equation}
\text{IoU} = \frac{|\mathcal{B}_{\text{pred}} \cap \mathcal{B}_{\text{gt}}|}{|\mathcal{B}_{\text{pred}} \cup \mathcal{B}_{\text{gt}}|},
\end{equation}
\begin{equation}
\text{NCD} = \frac{\sqrt{(x_{c}^{\text{pred}} - x_{c}^{\text{gt}})^{2} + (y_{c}^{\text{pred}} - y_{c}^{\text{gt}})^{2}}}{\sqrt{W_{\text{gt}}^{2} + H_{\text{gt}}^{2}}},
\end{equation}
\noindent where $\mathcal{B}_{\text{pred}}$ and $\mathcal{B}_{\text{gt}}$ represent the predicted and ground truth bounding boxes, respectively. The center coordinates of each bounding box are given by $(x_{c}^{\text{pred}}, y_{c}^{\text{pred}})$ and $(x_{c}^{\text{gt}}, y_{c}^{\text{gt}})$, while $W_{\text{gt}}$ and $H_{\text{gt}}$ denote the width and height of the ground truth box. The defect bounding boxes have been labeled manually and verified multiple times for consistency and spatial accuracy by cross-checking the annotations across independent review passes, ensuring reliable ground-truth localization for quantitative evaluation. This evaluation is discussed in Section \ref{subsec:defect_eval}. Sections \ref{subsec:ablation} and \ref{subsec:limitations} discuss ablation studies and highlight some limitations in the proposed framework, respectively.

\begin{table*}[p]
\centering
\renewcommand{\arraystretch}{2}
\caption{Qualitative comparisons between state-of-the-art AIRT dimensionality reduction techniques: TSR \cite{tsr}, PCA \cite{pca_2}, DAT \cite{dat}, 1D-DCAE-AIRT \cite{1d_cnn}, C-AET \cite{constrained_ae}, and the proposed AIRT-VLM adapter representation, under ambient and low-temperature ($-70^\circ$C) conditions.}
\begin{tabular}{|>{\centering\arraybackslash}m{2.5cm}|
                >{\centering\arraybackslash}m{3.2cm}|
                >{\centering\arraybackslash}m{3.2cm}|
                >{\centering\arraybackslash}m{3.2cm}|
                >{\centering\arraybackslash}m{3.2cm}|}
\hline
\textbf{Method} & \textbf{5 J} & \textbf{5 J ($-70^\circ$C)} & \textbf{15 J} & \textbf{15 J ($-70^\circ$C)} \\ \hline

\textbf{Raw} &
{\vspace{3pt}\includegraphics[width=3cm, height=2.5cm]{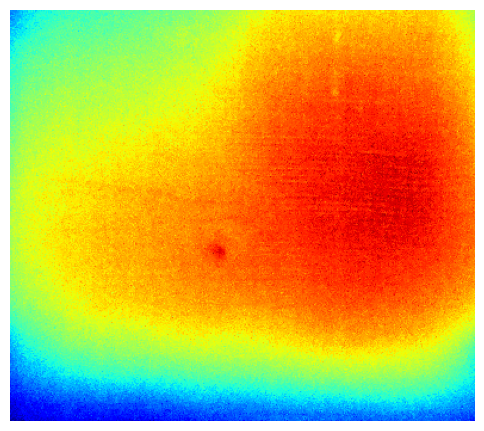}} &
{\vspace{3pt}\includegraphics[width=3cm, height=2.5cm]{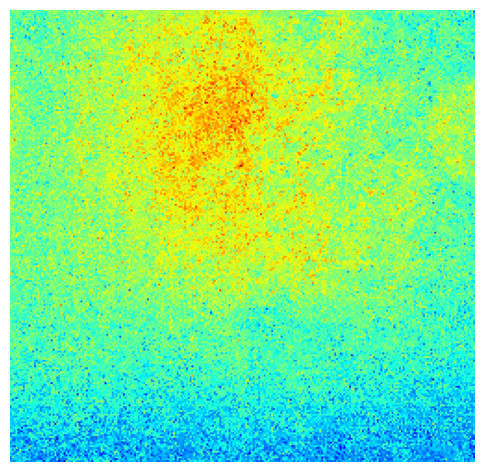}} &
{\vspace{3pt}\includegraphics[width=3cm, height=2.5cm]{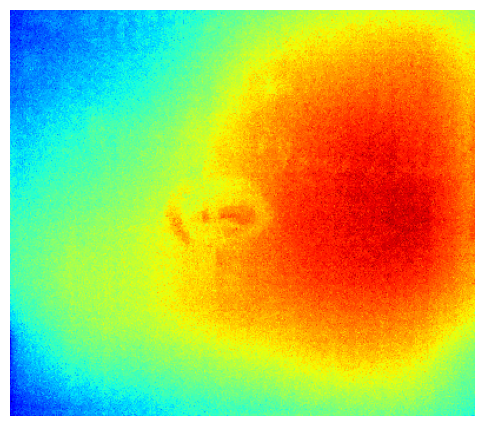}} &
{\vspace{3pt}\includegraphics[width=3cm, height=2.5cm]{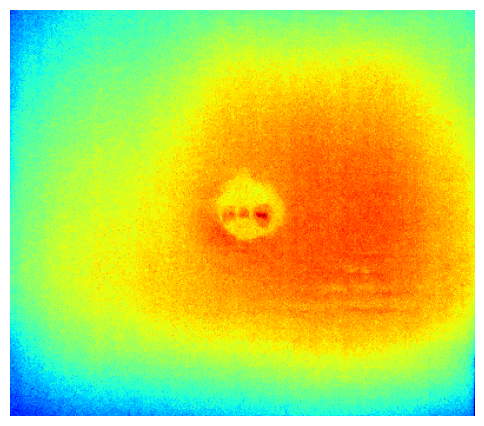}} \\ \hline

\textbf{TSR} &
{\vspace{3pt}\includegraphics[width=3cm, height=2.5cm]{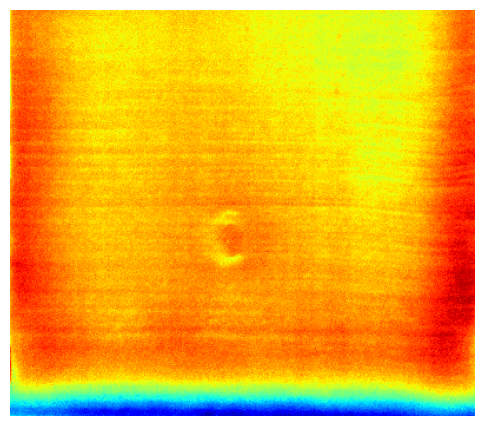}} &
{\vspace{3pt}\includegraphics[width=3cm, height=2.5cm]{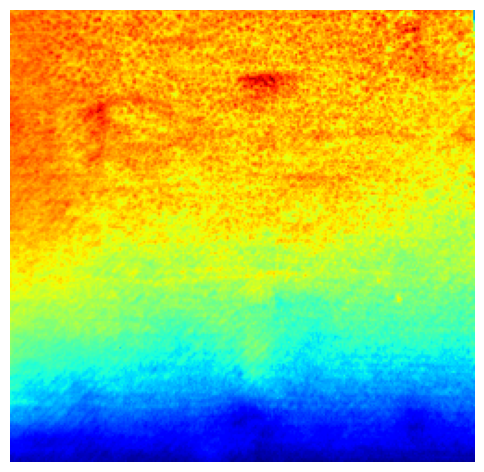}} &
{\vspace{3pt}\includegraphics[width=3cm, height=2.5cm]{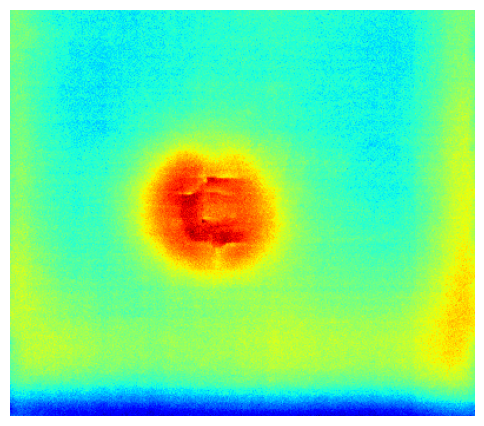}} &
{\vspace{3pt}\includegraphics[width=3cm, height=2.5cm]{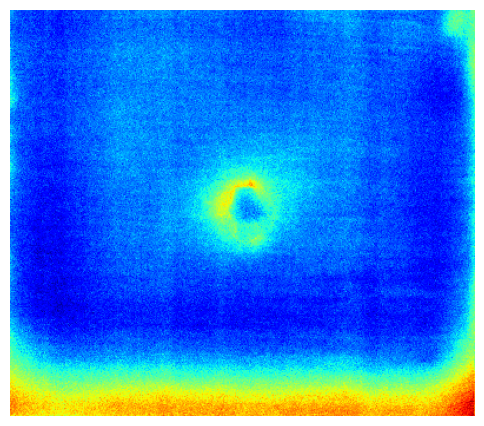}} \\ \hline

\textbf{PCA} &
{\vspace{3pt}\includegraphics[width=3cm, height=2.5cm]{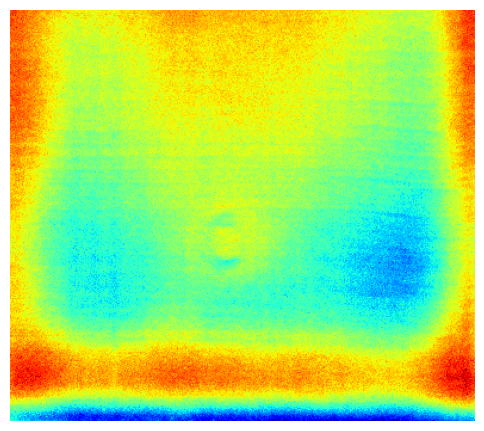}} &
{\vspace{3pt}\includegraphics[width=3cm, height=2.5cm]{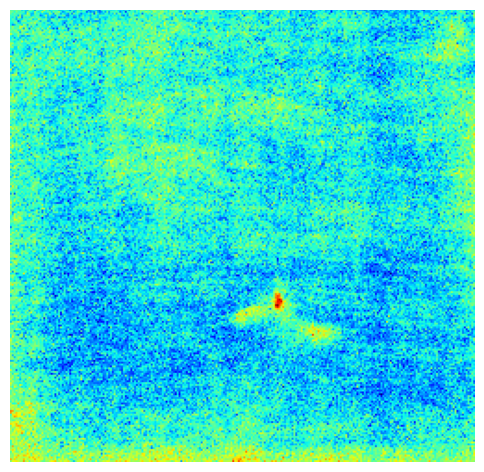}} &
{\vspace{3pt}\includegraphics[width=3cm, height=2.5cm]{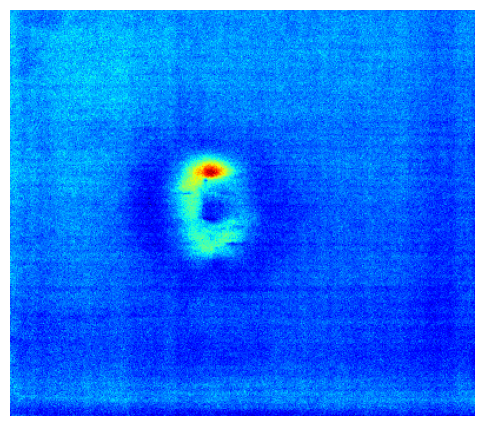}} &
{\vspace{3pt}\includegraphics[width=3cm, height=2.5cm]{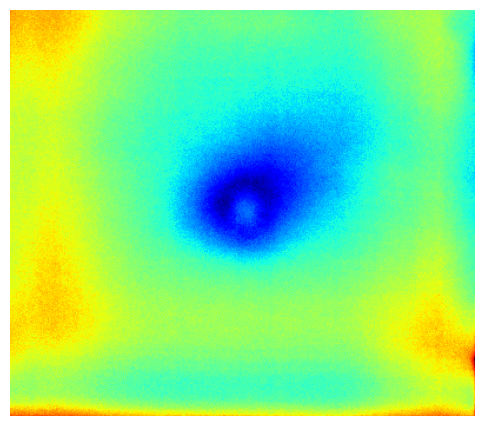}} \\ \hline

\textbf{DAT} &
{\vspace{3pt}\includegraphics[width=3cm, height=2.5cm]{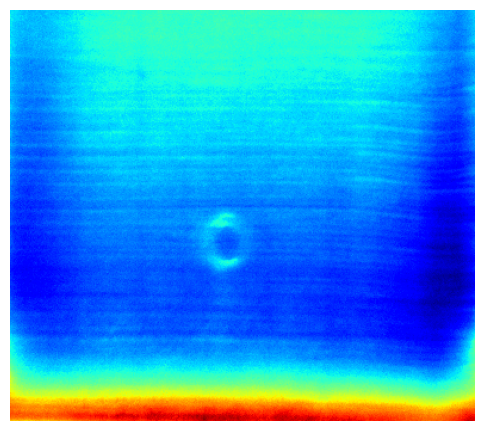}} &
{\vspace{3pt}\includegraphics[width=3cm, height=2.5cm]{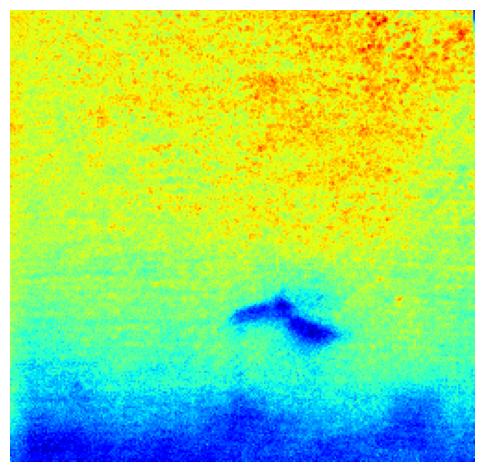}} &
{\vspace{3pt}\includegraphics[width=3cm, height=2.5cm]{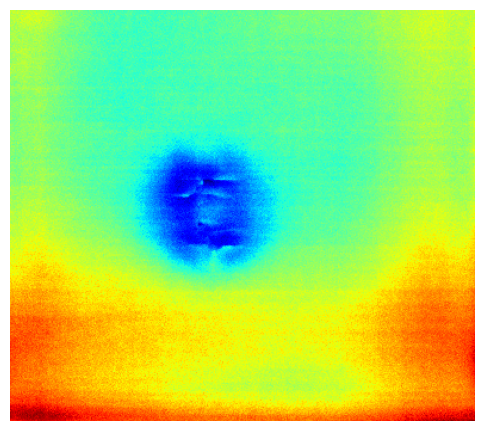}} &
{\vspace{3pt}\includegraphics[width=3cm, height=2.5cm]{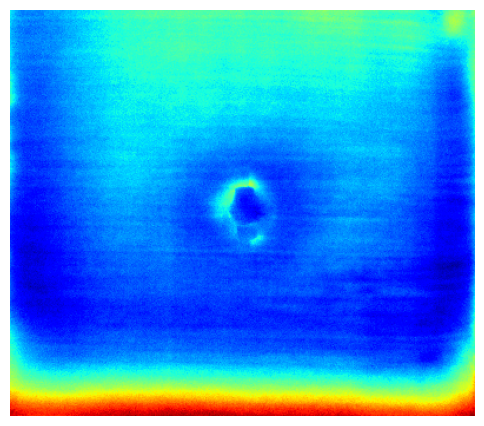}} \\ \hline

\textbf{1D-DCAE-AIRT} &
{\vspace{3pt}\includegraphics[width=3cm, height=2.5cm]{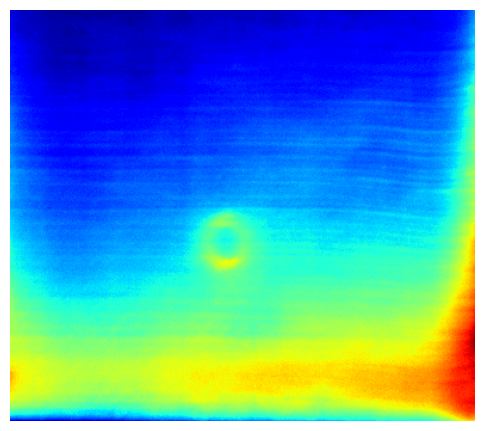}} &
{\vspace{3pt}\includegraphics[width=3cm, height=2.5cm]{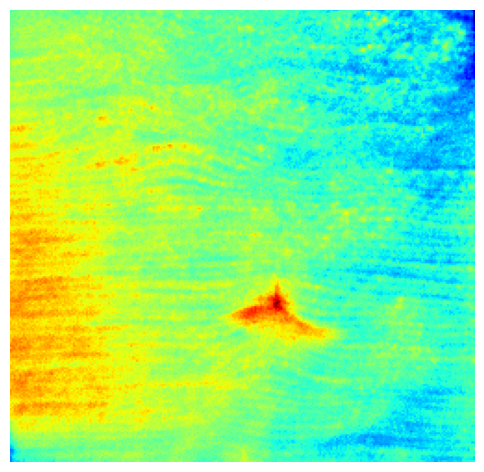}} &
{\vspace{3pt}\includegraphics[width=3cm, height=2.5cm]{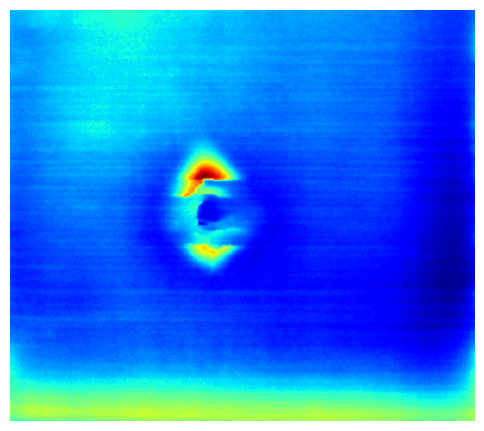}} &
{\vspace{3pt}\includegraphics[width=3cm, height=2.5cm]{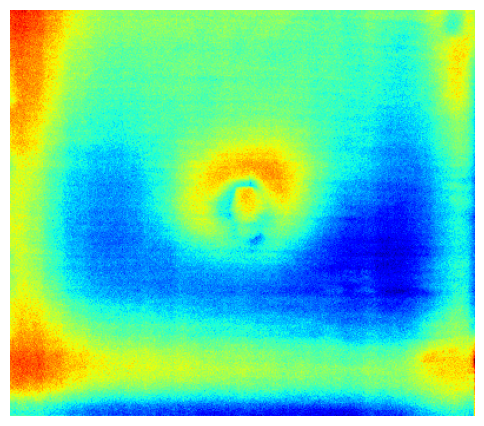}} \\ \hline

\textbf{C-AET} &
{\vspace{3pt}\includegraphics[width=3cm, height=2.5cm]{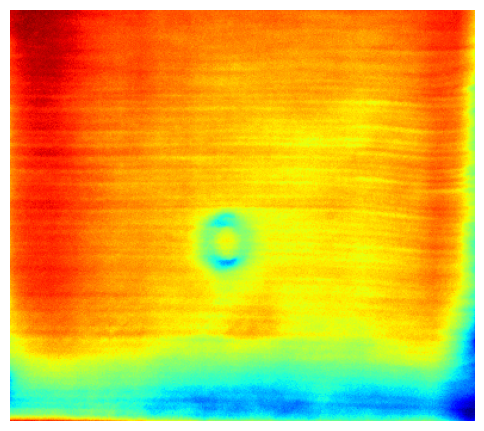}} &
{\vspace{3pt}\includegraphics[width=3cm, height=2.5cm]{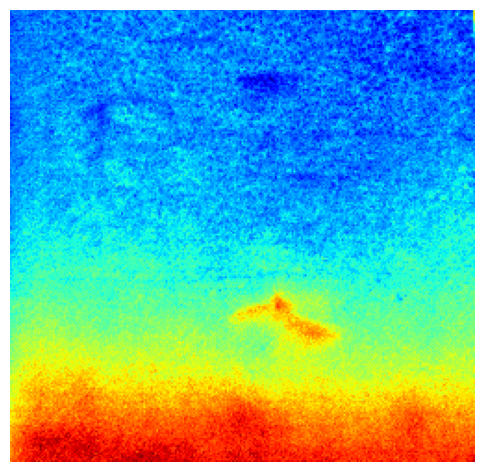}} &
{\vspace{3pt}\includegraphics[width=3cm, height=2.5cm]{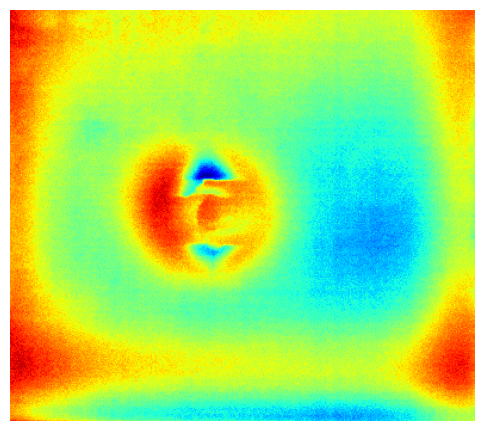}} &
{\vspace{3pt}\includegraphics[width=3cm, height=2.5cm]{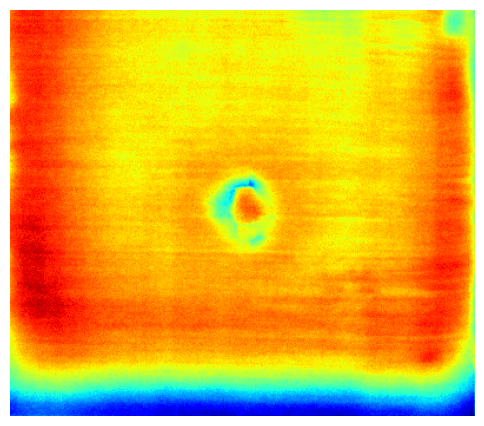}} \\ \hline

\textbf{Ours} &
{\vspace{3pt}\includegraphics[width=3cm, height=2.5cm]{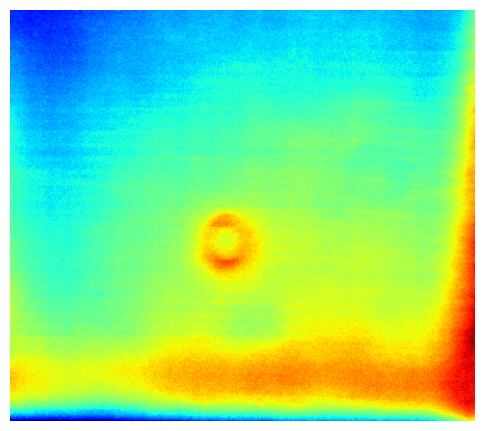}} &
{\vspace{3pt}\includegraphics[width=3cm, height=2.5cm]{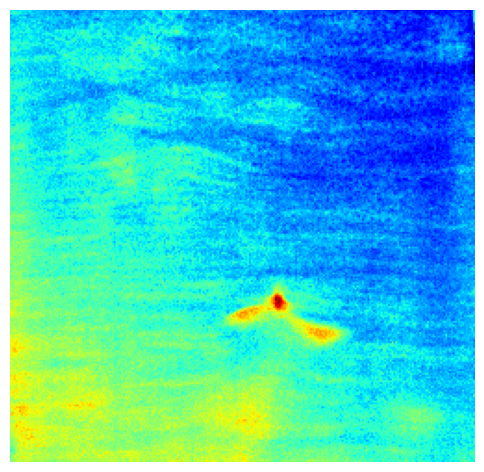}} &
{\vspace{3pt}\includegraphics[width=3cm, height=2.5cm]{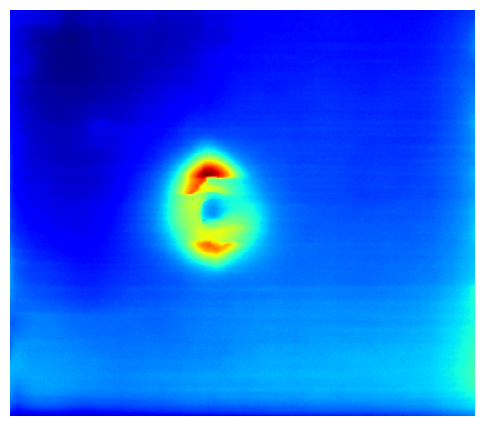}} &
{\vspace{3pt}\includegraphics[width=3cm, height=2.5cm]{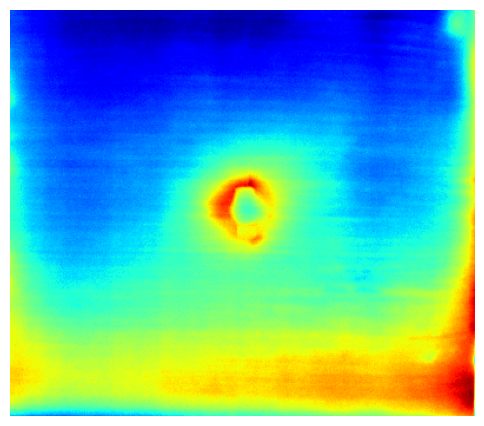}} \\ \hline

\end{tabular}
\label{table:qual_comparison}
\end{table*}

\subsection{AIRT-VLM Adapter Evaluation} \label{subsec:adapter_eval}
\begin{figure}[t]
\center
\includegraphics[keepaspectratio=true,scale=0.45, width=0.9\linewidth]{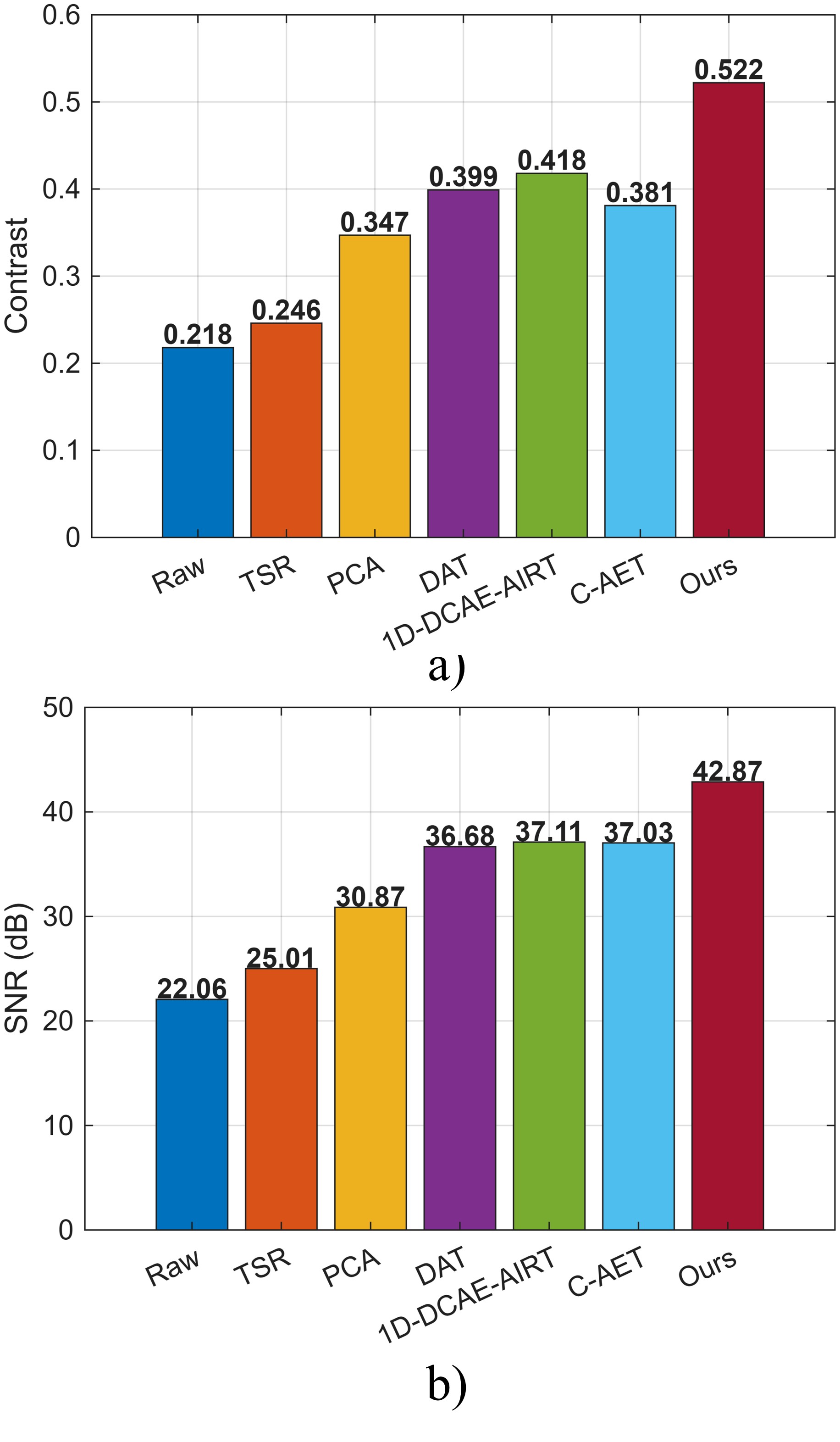}
\caption{Aggregate a) contrast and b) SNR on all 25 CFRP inspection sequences.}
\label{fig:contrast_snr}
\end{figure}

The purpose of the AIRT-VLM adapter is to mitigate the domain shift between thermal inspection data and the natural-image distributions on which VLMs are pretrained. For a thermal representation to be interpretable by a VLM, the defect patterns in the domain-aligned image must be visually distinct and sufficiently clear. This clarity reflects how effectively the AIRT-VLM adapter enhances defect-relevant signals while suppressing background noise and acquisition artifacts. To quantitatively assess this improvement, contrast and SNR metrics are used to evaluate the perceptual quality of the domain-aligned image. Table \ref{tab:contrast_snr} reports the contrast and SNR values obtained after applying the AIRT-VLM adapter and compares them against the raw thermogram as well as several state-of-the-art AIRT dimensionality reduction techniques, including PCA~\cite{pca_2}, TSR~\cite{tsr}, DAT~\cite{dat}, 1D-DCAE-AIRT~\cite{1d_cnn}, and C-AET~\cite{constrained_ae}. Fig.~\ref{fig:contrast_snr} summarizes the aggregated contrast and SNR performance across all methods, while Table~\ref{table:qual_comparison} presents qualitative comparisons illustrating the visual differences between the proposed AIRT-VLM representation and existing dimensionality reduction approaches.

The results in Table \ref{tab:contrast_snr} and Fig. \ref{fig:contrast_snr} show that the defect signal tends to increase with increasing impact energies across all sequences. This is expected since high impact energies create distinct defects that are easily identifiable compared to low impact defects. Nevertheless, the significant contrasts and SNR obtained for all defect classes are of the same order of magnitude. For instance, the proposed AIRT-VLM adapter increases the contrast by approximately 50\% and the SNR by 20 dB compared to the raw thermograms. On the other hand, the proposed framework outperforms traditional and learning-based AIRT dimensionality reduction methods. According to Table \ref{table:qual_comparison}, the visualizations highlight sharper defect boundaries, reduced halo artifacts, and superior suppression of background weave and non-uniform heating effects. Quantitatively, contrast improvements of up to 25\% and SNR gains exceeding 10 dB are observed compared to the strongest baseline, such as 1D-DCAE-AIRT. These improvements stem from the masked feature autoencoding strategy, which prevents trivial identity reconstruction and compels the network to focus on defect-relevant cues, the multi-level feature attention that amplifies salient channels across convolutional layers, and the self-attention block that captures long-range spatial and temporal dependencies. The obtained results also highlight that the AIRT-VLM adapter is capable of compressing a thermographic sequence to a single image, while exposing defect relevant features effectively. This is essential for reliable defect grounding, which is further discussed in the following section.

\begin{figure*}[!ht]
    \centering
    \includegraphics[keepaspectratio=true,scale=0.45, width=\linewidth]{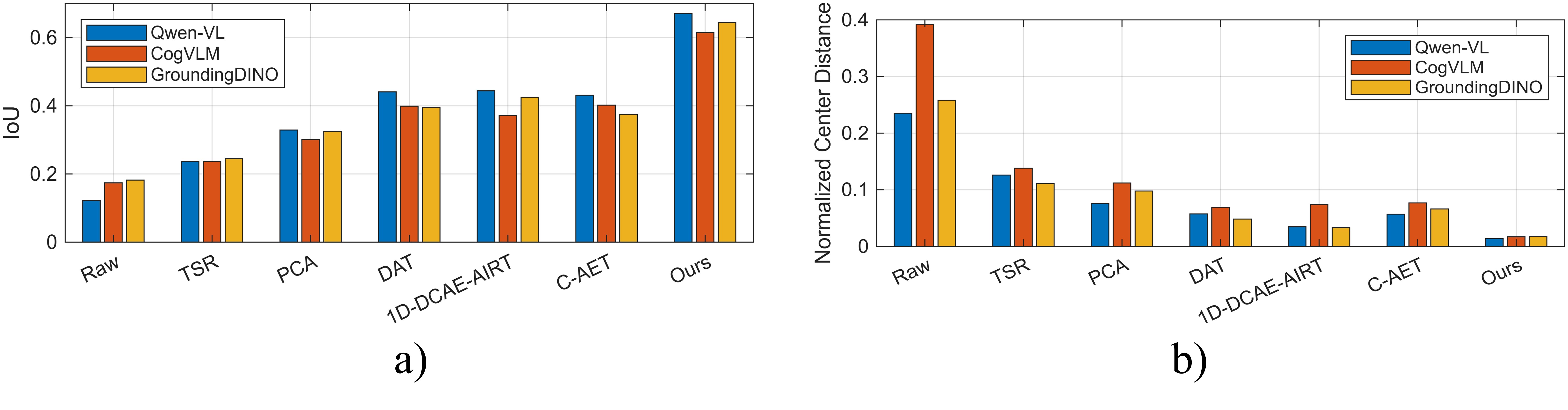}
    \caption{IoU and normalized center distance achieved by Qwen-VL, CogVLM, and GroundingDINO when coupled with AIRT-VLM adapter on all 25 CFRP inspection sequences. The performances of these models are also presented when coupled with state-of-the-art AIRT dimensionality reduction methods.}
    \label{fig:iou_dist}
\end{figure*}

\begin{table*}[!ht]
\centering
\renewcommand{\arraystretch}{2}
\caption{Defect detection results of Qwen-VL, CogVLM, and GroundingDINO coupled with the AIRT-VLM adapter, under ambient and low-temperature. The results show consistent defect detection performances across all samples.}
\resizebox{\linewidth}{!}{%
\begin{tabular}{|>{\centering\arraybackslash}m{2cm}|
                >{\centering\arraybackslash}m{3cm}|
                >{\centering\arraybackslash}m{3cm}|
                >{\centering\arraybackslash}m{3cm}|
                >{\centering\arraybackslash}m{3cm}|}
\hline
\textbf{Model} & \textbf{5 J} & \textbf{5 J ($-70^\circ$C)} & \textbf{15 J} & \textbf{15 J ($-70^\circ$C)} \\ \hline

\textbf{Qwen-VL}  &
{\vspace{3pt}\includegraphics[width=2.8cm,height=2.4cm]{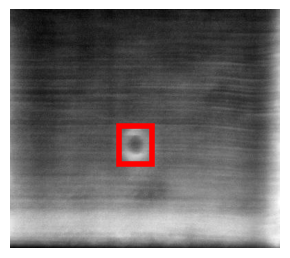}} &
{\vspace{3pt}\includegraphics[width=2.8cm,height=2.4cm]{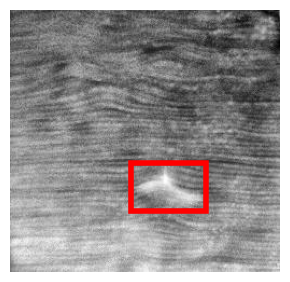}} &
{\vspace{3pt}\includegraphics[width=2.8cm,height=2.4cm]{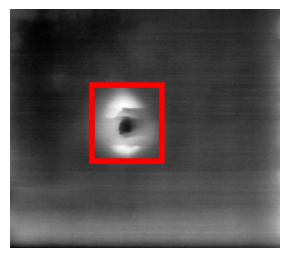}} &
{\vspace{3pt}\includegraphics[width=2.8cm,height=2.4cm]{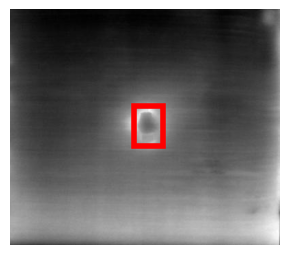}} \\ \hline

\textbf{CogVLM}  &
{\vspace{3pt}\includegraphics[width=2.8cm,height=2.4cm]{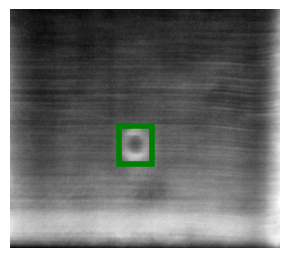}} &
{\vspace{3pt}\includegraphics[width=2.8cm,height=2.4cm]{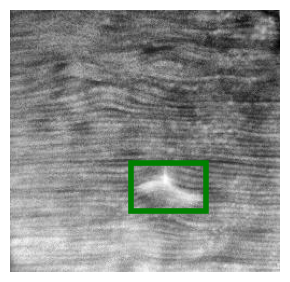}} &
{\vspace{3pt}\includegraphics[width=2.8cm,height=2.4cm]{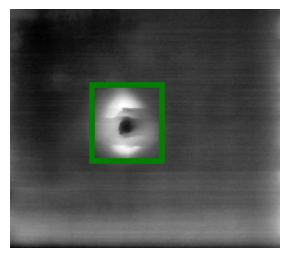}} &
{\vspace{3pt}\includegraphics[width=2.8cm,height=2.4cm]{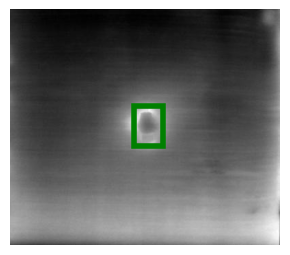}} \\ \hline

\textbf{GroundingDINO}  &
{\vspace{3pt}\includegraphics[width=2.8cm,height=2.4cm]{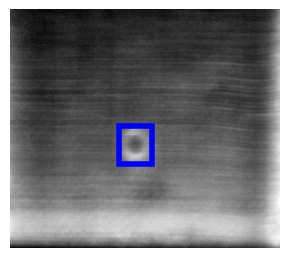}} &
{\vspace{3pt}\includegraphics[width=2.8cm,height=2.4cm]{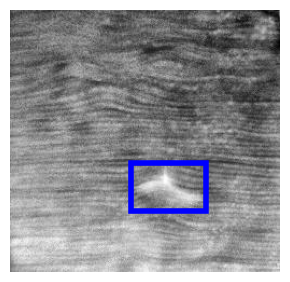}} &
{\vspace{3pt}\includegraphics[width=2.8cm,height=2.4cm]{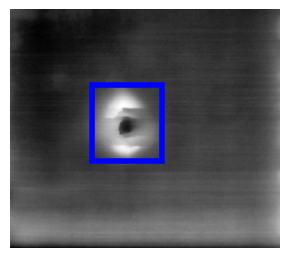}} &
{\vspace{3pt}\includegraphics[width=2.8cm,height=2.4cm]{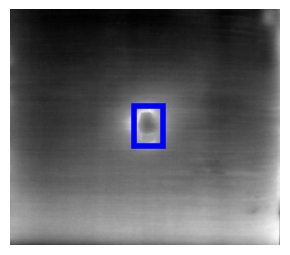}} \\ \hline

\end{tabular}
}
\label{table:vlm_qual_comparison}
\end{table*}

\subsection{Defect Detection Evaluation} \label{subsec:defect_eval}
The previous evaluation evaluated the AIRT-VLM adapter in terms of enhancing the clarity of the defects. Consequently, three different VLMs, Qwen-VL, CogVLM, and GroundingDINO, are evaluated on the AIRT-VLM adapter representation. The evaluation is based on the IoU and the normalized distance between the generated and ground truth bounding boxes. Note that the enhanced defect signal of the domain-aligned image results in a higher detection success rate and IoU. Table \ref{table:vlm_qual_comparison} shows the detected bounding boxes on two defective CFRP samples with impact damages at 5 J and 15 J. In addition, Fig. \ref{fig:iou_dist} displays the aggregate IoU and the normalized center distance of the VLMs on all sequences of each impact damage class. The figure also presents benchmarks, comparing the proposed framework against state-of-the-art dimensionality reduction methods coupled with the three VLMs for defect detection and grounding.

The results in Fig. \ref{fig:iou_dist} show that Qwen-VL, CogVLM, and GroundingDINO achieve IoUs higher than 60\%. when coupled with the AIRT-VLM adapter. Note that the size of the defects in the thermograms tends to be relatively small, approximately covering 5-10\% of the image size. As such, models that achieve IoUs exceeding 50\% tend to be within the acceptable range. Similarly, achieving a normalized center distance less than 0.05 Px provides emphasis on the model's grounding capabilities. Accordingly, coupling the VLMs with the AIRT-VLM adapter highlights strong zero-shot grounding capabilities with IoUs reaching to approximately 70\% and normalized center distance $\approx 0.015$. In addition, it is worth mentioning that all the models achieve consistent performances, demonstrating that the framework relies heavily on the domain-aligned input. If the performances are compared with the models' performance when coupled with the state-of-the-art AIRT dimensionality reduction techniques, the obtained IoUs do not exceed 50\%, demonstrating the importance of the AIRT-VLM adapter to generate domain-aligned images for stable and reliable subsurface defect grounding. These results show that defect detection in AIRT is attainable, where the subsurface defects are detected in a zero-shot manner, mitigating the challenge in time consuming preparation of AIRT training datasets.

\subsection{Ablation Studies} \label{subsec:ablation}
In addition to the previous evaluation routines, two studies are conducted on the proposed framework. The analysis in this section has been carried out on all 25 CFRP inspection seqeunces to ensure comprehensive and representative assessment. The first study studies the effect of pooling on latent image representation. As mentioned in Section \ref{sec:methods}, global average pooling is applied on the autoencoder latent space to generate a domain-aligned image representation for cognitive defect analysis. As such, the performance of the VLMs is analyzed when max pooling and PCA are applied on the latent images instead of average pooling. Table \ref{tab:ablation_1} outlines the contrast, SNR, IoU, and normalized center distance when utilizing the aforementioned pooling operations. Note that the IoU and normalized center distance are reported when using Qwen-VL. The results in Table \ref{tab:ablation_1} show that max pooling results in consistently lower defect detection performance across all metrics. This is because max pooling amplifies noise even in the presence of defective signals. On the other hand, average pooling exhibits similar performance to PCA. Both methods can be applied; however, the proposed framework opts for average pooling for its computational efficiency compared to PCA.

\begin{table}[h]
\centering
\caption{Quantified performances of average, max, and PCA pooling operations.}
\label{tab:ablation_1}
\resizebox{\linewidth}{!}{%
\begin{tabular}{l c c c}
\toprule
\textbf{Metric} & \textbf{Average Pooling} & \textbf{Max Pooling} & \textbf{PCA} \\
\midrule
\textbf{Contrast} & 0.522 & 0.471 & 0.547 \\
\textbf{SNR (dB)} & 42.87 & 39.18 & 42.41 \\
\textbf{IoU} & 0.691 & 0.539 & 0.701 \\
\textbf{Normalized Center Distance} & 0.0138 & 0.0378 & 0.0118 \\
\bottomrule
\end{tabular}%
}
\end{table}

In the second study, the elimination of the pooling operation is studied. The autoencoder generates $l=10$ latent images. Instead of reducing the latent space to a single domain-aligned image fed to the VLM, all $l=10$ images are fed to the VLMs, and then non-maximum suppression (NMS) is, consequently, applied to generate the defect bounding box. This acts as an ensemble operator to aggregate all bounding boxes from the inspection run on each latent image. Table \ref{tab:ablation_2} shows the performance of Qwen-VL when utilizing both aforementioned defect detection methods in terms of IoU, normalized center distance, and total execution time. The model performances show consistency across the two methods. Although the pooling operation tends to flatten the latent space to a single dimension, this comes at the cost of a 10-fold increase in execution time. Thus, the proposed multimodal defect analysis framework utilizes average pooling, which ensures comparable performance with less computational demands.

\subsection{Limitations} \label{subsec:limitations}
While the previous results show that zero-shot grounding of subsurface defects is attainable with the proposed framework, the work still has one limitation that opens the door to future research. Since the approach relies on dimensionality reduction and compresses the entire inspection sequence into a single image representation for VLMs, depth estimation of defects cannot be carried out. This is because the framework is designed to generate a domain-aligned image that resembles the type of data seen during VLM pretraining, which naturally leaves out the physics-based intuition contained in the full AIRT sequence. Another limitation; the framework cannot differentiate between defect types, such as delaminations, voids, or impact damage, which are common in real industrial settings. To differentiate between the aforementioned defect types, language-guided defect analysis cues need to be carried out in the spatiotemporal domain. At this stage, the method only identifies the presence of a defect. Future work will focus on fine-tuning VLMs to better capture the underlying physics of AIRT, enabling more generalizable defect analysis that includes identifying defect types and estimating defect depths to assess their severity. Yet, experimental validation across multiple impact energy levels—corresponding to defects with different geometrical characteristics—demonstrates that the proposed framework remains robust to variations in defect geometry and can reliably localize subsurface anomalies despite these structural differences.

\begin{table}[t]
\centering
\caption{Defect detection performance when the pooling operation is replaced with non-maximum suppression (NMS).}
\label{tab:ablation_2}
\resizebox{\linewidth}{!}{%
\begin{tabular}{l c c}
\toprule
\textbf{Metric} & \textbf{Average Pooling} & \textbf{NMS} \\
\midrule
\textbf{IoU}                        & 0.691  & 0.707  \\
\textbf{Normalized Center Distance} & 0.0138 & 0.0136 \\
\textbf{Execution Time (s)}         & 4.3    & 37.8   \\
\bottomrule
\end{tabular}%
}
\end{table}

\section{Conclusions} \label{sec:conc}

AI-driven active infrared thermography (AIRT) is increasingly adopted for automated inspection of composite materials; however, existing AI-based pipelines remain constrained by the need for large, labeled thermographic datasets and setup-specific training. To address these limitations, this work introduced a zero-shot cognitive defect analysis framework that integrates AIRT with off-the-shelf multimodal vision–language models (VLMs) through a lightweight AIRT–VLM adapter that produces domain-aligned thermal image representations. Experimental validation on 25 CFRP inspection sequences using Qwen-VL-Chat, CogVLM, and GroundingDINO demonstrates that the proposed method can localize subsurface defects without thermography-specific training or curated thermal datasets, achieving intersection-over-union (IoU) values of approximately 0.7 and normalized center distances (NCD) around 0.015. In practice, this enables accurate defect localization without curated thermal datasets, labeling procedures, or additional model retraining, resulting in a significant reduction in inspection setup time and overall analysis cost.

From an industrial perspective, the proposed framework removes the dataset bottleneck that currently limits the deployment of AI in thermography-based quality assurance, allowing rapid integration into existing inspection chains and providing repeatable, operator-independent defect localization. Overall, the proposed method bridges high-performing AIRT defect detection with the flexibility of multimodal AI, offering a viable route to scalable, training-free thermographic inspection. The combination of robust signal enhancement, zero-shot grounding, and minimal computational overhead positions this framework as a strong candidate for next-generation NDT systems, suitable for continuous monitoring, fast screening, and automated quality control of CFRP components. While effective for defect localization, future research will focus on fine-tuning VLMs with physics-informed objectives and leveraging richer temporal cues from AIRT sequences, enabling more generalizable cognitive defect analysis capable of identifying defect types and estimating their depth to assess severity.

\bibliographystyle{IEEEtran}
\bibliography{main.bib}

\end{document}